\documentclass[11pt]{article}
\usepackage[margin=1in]{geometry}
\usepackage[T1]{fontenc}
\usepackage{lmodern}
\usepackage{microtype}
\usepackage{booktabs}
\usepackage[section]{placeins}
\usepackage{tabularx}
\usepackage{amsmath}
\usepackage[numbers,sort&compress]{natbib}
\usepackage[hidelinks]{hyperref}
\usepackage{xcolor}

\title{Typical Angina, Reversed: A Long-Standing Documentation Error in the UCI Heart Disease Data}
\author{Tejas Tovinkere Pattabhi\\
University of the Cumberlands, Williamsburg, KY, USA\\
\texttt{tpattabhi65846@ucumberlands.edu}}
\date{Preprint, version 9, 25 November 2026}

\begin{document}
\maketitle

\begin{abstract}
The UCI Heart Disease data are a widely used machine-learning benchmark. We show that their chest-pain variable is stored in reverse of its documentation. At all four sites, the code documented as ``asymptomatic'' (4) marks patients with typical angina, and the code documented as ``typical angina'' (1) marks patients with almost none of the classical angina features (at Cleveland, no chest pain).

Two complementary checks establish this:
(i) the chest-pain table and regression coefficients of the original 1986 Cleveland study are reproduced only under the corrected reading;
(ii) in the Budapest, Switzerland and Long Beach files, 93--96\% of patients coded 4 have all three classical angina features.
A 1990 word-labeled Cleveland file already uses the documented labels, and Statlog (Heart), a Cleveland subset, stores the same codes. OpenML, PMLB and two highly voted Kaggle copies that name the categories use the same documented labels (about 400{,}000 downloads combined).

We audited a seeded random sample of 150 records from a pre-specified Scholar query for SHAP studies of these data mentioning asymptomatic chest pain (124 readable). Of 32 papers that state a direction for a named category, 25 (78\%; 95\% CI 61--89\%) state one inconsistent with the corrected category identities; 22 (69\%; 51--82\%; post hoc) are attributable to the documentation error, and nine of these explain it clinically. None of the 124 readable papers identifies the reversal. The primary codes were produced by Claude and reviewed by the author; the author's recoding of 52 papers (all 32 direction-stating papers and 20 others) agreed with them on 42 of the 50 that could be read ($\kappa=0.73$).

Conventional models are essentially unaffected, one reason the error went unnoticed. Interpretations that attach the documented names to model effects are reversed, and zero-shot language-model classifiers that read the labels are misled: with the documented labels, three open-weight models lose 0.12--0.33 in area under the receiver operating characteristic curve (AUC) given age, sex and chest-pain type.

We give the corrected mapping, evidence and code.
\end{abstract}

\section{Introduction}

The Heart Disease data of the UCI (University of California, Irvine) Machine Learning Repository \cite{uciheart} were collected at four centers in the 1980s for a study of probability-based diagnosis of coronary artery disease (CAD) \cite{detrano1989}. They have since become a standard teaching and benchmarking dataset. Their documentation codes chest-pain type (\texttt{cp}) as 1 = typical angina, 2 = atypical angina, 3 = non-anginal pain and 4 = asymptomatic.

Under this reading, the data say that patients without chest pain are the most likely to have CAD: 72.9\% of Cleveland patients coded ``asymptomatic'' have CAD, against 30.4\% of those coded ``typical angina''. 
This contradicts a basic and well-established gradient in cardiology. Typical angina is the presentation most predictive of CAD, and the absence of chest pain the least \cite{diamond1979}.

We show that the gradient is not a property of the patients but of the documentation: the codes are stored in reverse. Section~\ref{sec:evidence} establishes the reversal with two complementary checks, one for Cleveland and one for the other three sites, dates it and shows that Statlog (Heart) inherits it. Section~\ref{sec:spread} traces the documented labels into widely used copies. Section~\ref{sec:audit} measures how published work interprets the variable. Section~\ref{sec:consequences} examines what the error does and does not change.

The error matters for two reasons.
\begin{itemize}
\item Because it barely affects predictive accuracy, it could survive decades of benchmarking. A purely relabeled categorical variable leaves most models' predictions unchanged.
\item It does change interpretation, which is now a routine part of applied machine learning. SHAP (SHapley Additive exPlanations) values \cite{lundberg2017}, coefficients and association rules computed on these data attach clinical meaning to the documented labels. We find that published papers frequently report the backwards gradient as a clinical finding and sometimes explain it.
\end{itemize}

\textbf{Contributions.}
\begin{enumerate}
\item Evidence from two complementary sources, one for Cleveland and one for the other three sites, that \texttt{cp} is stored in reverse at all four sites, a dating of the error to 1990 or earlier, and evidence that Statlog (Heart) inherits it.
\item A trace of the documented labels through OpenML, PMLB (Penn Machine Learning Benchmarks) and Kaggle.
\item A pre-specified audit of a seeded random sample of 150 records from a targeted literature query.
\item An evaluation of the consequences for classical models, interpretations of their chest-pain effects, and language-model classifiers.
\item The corrected mapping and code to apply it.
\end{enumerate}

\section{Evidence that the coding is reversed}\label{sec:evidence}

\paragraph{Cleveland against the original study.}
Detrano et al.~\cite{detrano1986} report a study of 303 patients referred for coronary angiography at the Cleveland Clinic Foundation, the same number as the UCI Cleveland file.
\begin{itemize}
\item \textbf{Vessel counts:} the UCI outcome values 1--4 reproduce their Table~2 counts of one-, two- and three-vessel and left-main disease exactly (55, 36, 35, 13 patients).
\item \textbf{Chest pain by sex and disease status} (their Table~1): we matched the published row percentages to the UCI file in each of 16 cells (4 pain categories $\times$ sex $\times$ disease status). With the documented coding, 0 of 16 cells are consistent. With the corrected coding ($5-\texttt{cp}$), 15 of 16 are. Of all 24 possible code-to-category mappings, only this one matches more than 9 cells, so the table also supports the swap of the middle codes. The remaining cell's published percentage (27\% of 34) corresponds to no whole number of patients.
\item \textbf{Regression coefficients} (their Table~8; summarized in Table~\ref{tab:coef}): they report a logistic model of CAD on age, sex and chest pain (coded 1 = typical angina to 4 = no pain in their Methods): constant $-2.29$, age $0.07$, sex $-0.85$, pain $-1.10$. Refitting on the UCI file reproduces these values to two decimals only if pain is entered as $5-\texttt{cp}$, that is, if stored code 4 is the first category of their Table~1 (angina pectoris). With \texttt{cp} as stored, the constant becomes $-7.77$ and the pain coefficient changes sign to $+1.10$. In their larger seven-predictor model (297 patients), the corrected reading reproduces six slopes to two decimals and the seventh within 0.01. The constant differs ($-1.29$ against $-2.33$) by exactly the coefficient of calcified vessels (1.04), consistent with that variable being entered one unit higher than the UCI \texttt{ca} code. Recoding pain as $5-\texttt{cp}$ changes only the pain coefficient and the constant, so the decisive evidence is the sign of the pain coefficient and the constant of the three-predictor model.
\end{itemize}

\begin{table}[t]
\centering
\caption{Logistic regression of CAD on age, sex and chest pain in the Cleveland data (303 patients): coefficients published by Detrano et al.~\cite{detrano1986} (Table~8, subset~1) and refitted on the UCI file with \texttt{cp} as stored and corrected ($5-\texttt{cp}$).}\label{tab:coef}
\begin{tabular}{lcccc}
\toprule
 & Constant & Age & Sex & Pain\\
\midrule
Published & $-2.29$ & $0.07$ & $-0.85$ & $-1.10$\\
Refit, \texttt{cp} as stored & $-7.77$ & $0.07$ & $-0.85$ & $+1.10$\\
Refit, \texttt{cp} corrected ($5-\texttt{cp}$) & $-2.29$ & $0.07$ & $-0.85$ & $-1.10$\\
\bottomrule
\end{tabular}
\end{table}

\paragraph{Budapest, Switzerland and Long Beach against their own raw records.}
The 76-attribute raw files for these three sites record the three features that define typical angina \cite{diamond1983}: substernal location, provocation by exertion, and relief by rest. The UCI documentation lists them as attributes 5--7, and their sum as attribute 8. Table~\ref{tab:raw} cross-tabulates them against \texttt{cp}.
\begin{itemize}
\item Patients coded 4 almost always have all three features.
\item Patients coded 1 almost never have any.
\item At each site, the mean number of features falls monotonically from code 4 to code 1.
\end{itemize}
This matches the three-feature definitions of typical, atypical and non-anginal pain, in reverse, and code 1 is consistent with no chest pain. Exercise-induced angina, recorded separately in the processed files, is also most frequent in code 4 at every site (51--71\%, against 0--55\% in the other codes). 
In the Cleveland raw file these fields are missing ($-9$) in all 282 parsable records, and the archive's own \texttt{WARNING} file notes that this file is corrupted.

\begin{table}[t]
\centering
\caption{Classical angina features (substernal, exertional, relieved by rest) by stored chest-pain code, in the raw records of three UCI sites. Each cell gives the number of patients with 0, 1, 2 or 3 of the features, among patients with all three recorded. Full tables for all four codes are in the repository.}\label{tab:raw}
\begin{tabular}{lcccc}
\toprule
Site / stored code & 0 features & 1 & 2 & 3 features\\
\midrule
Budapest, code 1 (``typical angina'') & 10 & 0 & 0 & 1\\
Budapest, code 4 (``asymptomatic'') & 0 & 0 & 5 & 118\\
Switzerland, code 1 & 4 & 0 & 0 & 0\\
Switzerland, code 4 & 1 & 3 & 3 & 91\\
Long Beach, code 1 & 8 & 0 & 0 & 0\\
Long Beach, code 4 & 0 & 0 & 5 & 126\\
\bottomrule
\end{tabular}
\end{table}

\paragraph{The documented labels were in use by 1990.}
The UCI archive includes \texttt{cleve.mod}, a word-labeled version of the Cleveland data prepared by John Gennari in March 1990. Joined to the processed file on age, resting blood pressure, cholesterol and maximum heart rate, its 303 rows (302 distinct patients; the file repeats one record and omits one) map code 1 to \texttt{angina}, 2 to \texttt{abnang}, 3 to \texttt{notang} and 4 to \texttt{asympt}. The documented labels were therefore already in use in 1990. 
Its consequence is visible in the same file: 73\% of patients labeled \texttt{asympt} are \texttt{sick}.

\paragraph{Statlog (Heart).}
The Statlog (Heart) dataset \cite{ucistatlog} (270 patients) matches 270 Cleveland rows uniquely, with identical chest-pain codes and outcomes in every row. It inherits the error. 

\paragraph{Correction.}
The stored patient values are internally consistent; what is wrong is the category documentation. For all four UCI files and Statlog, the clinically correct label of stored code $k$ is the documented label of $5-k$: 4 = typical angina, 3 = atypical angina, 2 = non-anginal pain, 1 = asymptomatic.

The middle two codes swap as well. The raw features support this at Budapest and Long Beach, where code 3 patients mostly have two features and code 2 patients mostly one. 
Under the corrected labels, Cleveland CAD prevalence is 72.9\% for typical angina (144 patients), 21\% for atypical angina (86), 18\% for non-anginal pain (50) and 30.4\% for asymptomatic patients (23). 
At all four sites, asymptomatic patients have higher prevalence than those with non-anginal pain, although the groups are small (4 to 23 patients). This may reflect referral selection: patients without chest pain are likely sent for angiography only when other findings suggest disease. The identification of code 1 rests on Detrano's Table~1 and the raw features, not on prevalence.

\section{Where the documented labels spread}\label{sec:spread}

We downloaded widely used copies (OpenML, PMLB and the most-voted Kaggle copies) and matched their rows to the UCI files (17--18~Nov 2026). Table~\ref{tab:spread} summarizes the result. Every copy we checked whose page or metadata names the categories uses the documented labels: it keeps the UCI codes and documentation, writes the labels out as words, or recodes the values and documents the recoded labels in the same order. PMLB (Penn Machine Learning Benchmarks) \cite{pmlb}, for example, stores UCI$-1$ and documents 0 as typical angina. One copy differs: the Kaggle copy by D.~Lapp \cite{kagglelapp}, the most downloaded of those we checked, swaps the end codes (0 = UCI 4, 3 = UCI 1), keeps the middle codes in order, does not state the labels, and reverses the outcome. Its 1{,}025 rows contain only 302 unique Cleveland patients. Under the usual 0-based reading (0 = typical angina), its end codes are correct; the copy looks backwards only because the outcome is reversed (25\% of code-0 patients have target 1). Some users instead read code 0 as asymptomatic, the documented UCI label of code 4, which carries the documentation error into the recoded copy. A 303-row Kaggle copy \cite{kagglenaresh} has the same coding and reversed outcome.

\begin{table}[t]
\centering
\small
\caption{Coding of chest-pain type in widely used copies (coding checked 17--18 Nov 2026; usage counts read from the pages on 18 Nov 2026). ``Wrong'' means the copy's own labels give the documented reading.}\label{tab:spread}
\begin{tabularx}{\linewidth}{lXl}
\toprule
Copy & Chest-pain coding & Reach\\
\midrule
UCI repository page \cite{uciheart} & documents 1 = typical angina; wrong & 1.26M views\\
Kaggle, fedesoriano \cite{kagglefedesoriano} (918 rows, 4 sites + Statlog) & words; ``TA'' = UCI 1; wrong & 290K downloads\\
Kaggle, D.~Lapp \cite{kagglelapp} (1{,}025 rows) & 0 = UCI 4, 3 = UCI 1; labels not stated; outcome reversed & 386K downloads\\
Kaggle, redwankarimsony \cite{kaggleredwan} (920 rows) & words; ``typical angina'' = UCI 1; wrong & 112K downloads\\
OpenML \texttt{heart-c}, \texttt{heart-h} \cite{openml} & words; \texttt{typ\_angina} = UCI 1; wrong & ---\\
PMLB \texttt{heart\_disease\_*} \cite{pmlb} & UCI$-1$, documented 0 = typical angina; wrong & ---\\
\bottomrule
\end{tabularx}
\end{table}

\section{How published work reads the variable}\label{sec:audit}

\paragraph{Protocol.}
The protocol was fixed before any full text was read; later changes are logged with dates in the accompanying repository.
\begin{itemize}
\item \textbf{Frame:} Google Scholar records from 2010 onward matching \textit{``chest pain'' ``asymptomatic'' ``Cleveland'' ``heart disease'' ``SHAP''} (269 records; 230 eligible after removing off-topic records and duplicates).
\item \textbf{Sample:} the first 150 records of a seeded random order (seed 20261117) of the 230 eligible records.
\item \textbf{Inclusion:} a paper is included if its full text states a direction for a named chest-pain category, from its own data or model. Importance without a sign, numeric-only statements and background text were not counted.
\item \textbf{Coding:} for each included paper we recorded the copy used, the verbatim claim, whether the stated direction is clinically correct, and whether the paper explains the direction clinically. Claude (Anthropic) produced the primary codes, and the author reviewed them.
\item \textbf{Repeat coding:} about a week later, the author recoded a shuffled packet containing all 32 papers originally coded as direction-stating and 20 others, without consulting the per-paper primary codes. An automated extraction step supplied text passages. The author checked source material where available, including 13 publisher PDFs (11 through university library access and 2 open access), and used tables and figures when they established the category and direction. Two fresh-context reviews with OpenAI Codex, also without the primary codes, then checked the recoding against the coding rules. The first review changed the direction code for two papers and the rationalization code for one. Coding was frozen after the second review. Raw agreement and Cohen's $\kappa$ were calculated on the 50 papers coded in both passes.
\item \textbf{Not reported:} the pre-specified citation-weighted totals (descriptive) were not computed.
\item \textbf{Access:} inaccessible papers were counted, not replaced. Full texts came from open access, the author's university library and open repository copies. No access control was circumvented: where a publisher showed a browser check, the author cleared it by hand or the paper was counted as inaccessible.
\end{itemize}

\paragraph{Results.}
124 of the 150 full texts were readable.
\begin{itemize}
\item \textbf{Direction-stating papers:} 32 state a direction for a named category. In 25 of the 32 (78\%; 95\% Wilson interval \cite{wilson1927} 61--89\%), the pre-specified outcome, the stated direction is inconsistent with the corrected category identities. Most report the category labeled asymptomatic as highest risk, or typical angina as lowest. Three of these are inconsistent for other reasons: two rest on Kaggle copies whose outcome is reversed, and one uses its own mapping. Excluding them, a secondary refinement defined after coding, leaves 22 of 32 (69\%; 51--82\%) attributable to the documented labels. Statements based on recoded Kaggle copies count as attributable only when the inconsistency comes from naming a category through the documented UCI label (for example, code 0 read as asymptomatic), not from the copy's reversed outcome.
\item \textbf{Sensitivity analysis} (criteria partly defined after coding): excluding hedged, illustrative, single-patient, mixed and inconsistently mapped statements leaves 18 of 27 attributable to the documentation (67\%; 48--81\%), or 20 of 27 counting all inconsistencies (74\%; 55--87\%). Applying the same criteria to the correct-sounding statements removes one more, giving 18 of 26 (69\%; 50--83\%).
\item \textbf{Typical wording:} ``Patients with asymptomatic chest pain (ASY) are about six times more likely to suffer heart disease than those with Atypical Angina\ldots'' \cite{baghdadi2023}.
\item \textbf{Rationalized errors:} nine of the 22 also explain the inconsistent direction clinically, for example:
  \begin{itemize}
  \item ``This may seem counterintuitive, but asymptomatic chest pain is actually the most severe out of the four types'' \cite{dave2020};
  \item ``\ldots aligning with clinical findings that many severe cardiovascular conditions present with minimal symptoms'' \cite{harrou2025}.
  \end{itemize}
\item \textbf{The seven correct-sounding statements:} six, on the available evidence, do not reflect the true coding; the seventh cannot be assessed.
  \begin{itemize}
  \item Three contradict the paper's own coding table or its own data.
  \item Two appear, from their counts or correlations, to use a 303-row Kaggle copy that shares the Lapp copy's coding and reversed outcome, so the two errors cancel.
  \item One maps codes inconsistently.
  \item One does not identify its copy.
  \end{itemize}
\item \textbf{Copies behind the 25 inconsistent statements:} UCI files or copies that keep the UCI codes or labels, including 1{,}190-row Kaggle merges (12); the fedesoriano copy (5); recoded Kaggle copies (6; three of these are inconsistent for other reasons); and copies the papers do not identify (2).
\end{itemize}
None of the 124 readable papers identifies the reversal, including a 2026 leakage audit devoted to these data \cite{hussain2026}.
The 32 direction-stating papers, with venues, copies, verbatim claims and codings, are listed in \texttt{evidence/literature-audit/direction-stating-32.tsv} in the repository.

\paragraph{Repeat coding.}
The recoding covered 50 of the 52 papers; two full texts remained behind paywalls. Agreement between the primary coding and the final recoding on the three-way outcome (no direction, clinically inconsistent, clinically consistent) was 42 of 50 ($\kappa=0.73$; 44 of 50 before the first review). The two codings agreed on whether a paper stated a direction for 43 of 50 papers ($\kappa=0.72$) and on clinical consistency for 24 of the 25 papers that both identified as stating a direction ($\kappa=0.86$). On the same 50 papers, the primary coding found 24 inconsistent statements among 30 direction-stating papers (80\%), and the recoding found 21 among 27 (78\%). 

\section{Consequences}\label{sec:consequences}

\paragraph{Discrimination of conventional models is essentially unchanged.}
On Cleveland (303 patients, five-fold cross-validation), we fitted the same models with \texttt{cp} as stored and with \texttt{cp} relabeled:
\begin{itemize}
\item \textbf{Logistic regression,} with \texttt{cp} ordinal or one-hot: identical predictions (area under the receiver operating characteristic curve, AUC, 0.903 and 0.911 under both codings).
\item \textbf{Random forests and gradient boosting:} AUC differs by at most 0.002. Individual predictions differ slightly (by up to 0.07 and 0.15), because relabeling reorders the one-hot columns and so changes the randomized splits.
\end{itemize}
This is one reason benchmarking did not reveal the error. 

\paragraph{Interpretations of chest-pain effects are reversed.}
The same one-hot logistic model (scikit-learn defaults, L2 penalty; missing values median-imputed), fitted to all 303 patients with standardized features, assigns the largest chest-pain coefficient ($+0.48$) to the category labeled asymptomatic and a negative one ($-0.25$) to the category labeled typical angina. An analyst using the documented labels could read this as ``no chest pain raises risk''. The audit in Section~\ref{sec:audit} shows that published work does so.

\paragraph{Language-model classifiers are affected.}
Tabular methods based on large language models (LLMs) serialize each record as text and rely on the model's prior knowledge \cite{hegselmann2022}. For these methods the label is an input, not just an annotation.

We ran an exploratory probe on Cleveland whose protocol was fixed before any model was run:
\begin{itemize}
\item \textbf{Models:} three open-weight instruction models (Qwen2.5-7B \cite{qwen25}, Llama-3.1-8B \cite{llama3} and Phi-3.5-mini \cite{phi3}, 4-bit), zero-shot.
\item \textbf{Score:} the probability of ``Yes'' relative to ``No'' for the first answer token of a yes/no question about CAD.
\item \textbf{Templates:} each patient was described with all 13 predictors (\emph{full}) or with age, sex and chest pain only (\emph{minimal}).
\item \textbf{Conditions:} documented labels (D), corrected labels (C), or no chest-pain line.
\end{itemize}

Table~\ref{tab:llm} gives the results.
\begin{itemize}
\item Corrected labels improved log loss in all six model--template pairs.
\item With the minimal template, corrected labels improved AUC by 0.12--0.33 in all three models. With the documented labels, Qwen's AUC fell to 0.458, at or below chance, and Qwen and Llama both scored worse than with chest pain omitted.
\item With the full template, AUC improved in two of three models (Qwen $+0.016$, Llama $+0.077$; intervals exclude 0). Phi was unchanged.
\item For Llama with the minimal template, mean predicted risk rose from code 1 to code 4 under the corrected labels (0.02 to 0.29) and fell under the documented labels (0.29 to 0.02). Qwen and Phi showed the same reversal of the code-1/code-4 ordering with the minimal template, although their mean probabilities by code stayed below 0.001.
\end{itemize}
Phi (both templates) and Qwen (minimal template) almost never favored ``Yes'': no patient's P(Yes) exceeded 0.015 for Phi or 0.003 for Qwen with the minimal template, under any condition. Their AUC differences therefore reflect the ranking of very small probabilities, and their Brier-score differences are negligible ($|\Delta\text{Brier}| \le 0.0003$; for Phi with the full template the interval includes 0). For Llama (both templates) and Qwen (full template), the Brier score improved with the corrected labels ($\Delta$Brier $-0.022$ to $-0.144$; intervals exclude 0).

Llama discriminated poorly on the full template under both labelings (AUC 0.51--0.59), so its full-template gain should be read with caution. We set a threshold in advance for scaling the probe up to larger models: an AUC gain of at least 0.02 on the full template in two models, or a reversed risk gradient in two models. On the full template, neither criterion holds in two models. The AUC gain reached 0.02 only for Llama (Qwen $+0.016$, Phi $+0.002$), and the ordering of codes 1 and 4 reversed only for Llama. With the minimal template, the ordering of codes 1 and 4 reversed in all three models, but for Qwen and Phi the mean probabilities being compared are all below 0.001. In no model or template does mean P(Yes) rise steadily across codes 1, 2, 3 and 4 under the corrected labels; the reversal holds only when code 1 is compared with code 4. The protocol does not say which template the gradient criterion refers to, or whether ``rises'' means from code 1 to code 4 or at every step. The threshold is therefore met only on the reading that uses the minimal template and compares the end codes. We report the full-template effect as modest.

\textbf{Protocol deviation.} The protocol specified a replication on the UCI Hungarian data, to be reported if the primary hypothesis held. The primary hypothesis held for Qwen and Llama. The scale-up threshold is met only on the reading described above (minimal template, end codes), and on the full template it is not met. Neither the Hungarian replication nor the scale-up to larger models was run. We record both as deviations from the protocol. The pre-specified Brier-score differences are summarized above and listed in the repository with the per-patient scores. The runs used Kaggle rather than Colab GPUs.

\begin{table}[t]
\centering
\footnotesize
\setlength{\tabcolsep}{4pt}
\caption{LLM probe on Cleveland (303 patients). AUC under documented (D) and corrected (C) labels, and differences C$-$D with paired-bootstrap 95\% intervals (2{,}000 resamples).}\label{tab:llm}
\begin{tabular}{llccccc}
\toprule
Model & Template & AUC D & AUC C & AUC no cp & $\Delta$AUC [95\% interval] & $\Delta$ log loss [95\% interval]\\
\midrule
Qwen2.5-7B & full & 0.844 & 0.861 & 0.859 & $+0.016$ [0.004, 0.030] & $-0.128$ [$-0.215$, $-0.042$]\\
Qwen2.5-7B & minimal & 0.458 & 0.789 & 0.637 & $+0.331$ [0.240, 0.425] & $-1.788$ [$-2.108$, $-1.487$]\\
Llama-3.1-8B & full & 0.512 & 0.589 & 0.477 & $+0.077$ [0.043, 0.114] & $-0.110$ [$-0.139$, $-0.082$]\\
Llama-3.1-8B & minimal & 0.542 & 0.779 & 0.626 & $+0.238$ [0.145, 0.338] & $-0.846$ [$-1.003$, $-0.700$]\\
Phi-3.5-mini & full & 0.855 & 0.856 & 0.837 & $+0.002$ [$-0.012$, 0.016] & $-0.107$ [$-0.148$, $-0.067$]\\
Phi-3.5-mini & minimal & 0.673 & 0.795 & 0.638 & $+0.122$ [0.056, 0.191] & $-0.955$ [$-1.127$, $-0.798$]\\
\bottomrule
\end{tabular}
\end{table}

\section{Related work}

\begin{itemize}
\item \textbf{Label and documentation errors in benchmark data:} label errors in widely used test sets alter model rankings \cite{northcutt2021}. Idiosyncrasies of the UCI Adult data, uncovered by reconstructing its census provenance, motivated replacement datasets \cite{ding2021}. Datasheets propose documentation that would surface such problems \cite{gebru2021}, and studies of data cascades show that data issues compound silently downstream \cite{sambasivan2021}.
\item \textbf{Provenance errors in these data:} Ogu et al.~\cite{ogu2026} report that a widely shared Kaggle copy of the Cleveland data reverses the outcome and recodes chest pain. They treat the UCI documentation as correct.
\item \textbf{Backwards rules learned from data:} a well-known clinical example of an explainable model learning a medically backwards rule from real data is the asthma--pneumonia case of Caruana et al.~\cite{caruana2015}. There the rule reflected a real feature of care. Here the rule reflects a documentation error.
\end{itemize}
To our knowledge, no prior work reports that the UCI files store chest-pain type in reverse of their documentation. We searched the web, Google Scholar, Semantic Scholar, IEEE Xplore, Kaggle discussions, GitHub issues and the UCI page between 16 and 18 Nov 2026.

\section{Discussion}

\paragraph{Recommendations.}
\begin{itemize}
\item \textbf{Maintainers} of the UCI page, OpenML, PMLB and the Kaggle copies should correct the chest-pain labels and add a note to existing versions.
\item \textbf{Users} should map stored code $k$ to the documented label of $5-k$ for the UCI files and Statlog. For copies coded 0--3, they should check which transformation applies (Table~\ref{tab:spread}).
\item \textbf{Papers that interpret chest-pain effects} from these data should re-examine those interpretations. Most conventional predictive results are unaffected; text-based models may not be.
\end{itemize}
More generally, a benchmark variable whose effect contradicts settled domain knowledge is a reason to check the data before explaining the finding.

\paragraph{Limitations.}
\begin{itemize}
\item \textbf{Cleveland evidence:} the Cleveland raw file lacks the pain features, so the evidence for Cleveland rests on the original study alone. The 1990 file dates the reading, and Statlog only shows inheritance.
\item \textbf{Literature frame:} the frame is one Scholar query that requires the words asymptomatic and SHAP. Our rates describe direction-stating papers in this frame, not the whole literature. Because the query requires the word asymptomatic, papers in the frame are probably more likely to state a direction for that category than papers in general, so the 78\% may be higher than in the wider literature.
\item \textbf{Coding and recoding:} the primary codes were produced by Claude and reviewed by the author; the recoding was done by the author, with review by OpenAI Codex. The author had reviewed the primary codes about a week before recoding and knew the audit's overall results, so the agreement figures describe a human check of AI coding. The 22 of 32 attributable statements and the sensitivity counts also depend on categories defined after coding. The agreement estimates apply to the 50 accessible papers in a packet enriched for papers originally coded as direction-stating. The automatically extracted passages could omit relevant context or claims in tables and figures, and only some of those materials were checked directly. Attribution and sensitivity comparisons reused the primary coder's exclusions. No coder unconnected with the project has recoded the audit.
\item \textbf{Unread papers:} 17\% of sampled papers could not be read.
\item \textbf{Text-only extraction:} in the primary audit, claims shown only in figures were not captured, so the counts of inconsistent statements are lower bounds. The percentages could move either way, because the number of direction-stating papers could also change.
\item \textbf{LLM probe:} small quantized models, one prompt format, zero-shot only, and one site. The findings may not hold for few-shot or fine-tuned methods, where the data can override the label text. The pre-specified Hungarian replication was not run.
\item \textbf{Copies checked:} Kaggle copies were selected from a list sorted by votes, not downloads; other popular copies may differ.
\end{itemize}

\section{Conclusion}

The UCI Heart Disease chest-pain variable has been documented in reverse since at least 1990. The error has little effect on discrimination, which helps explain why it persisted. It reverses interpretations of chest-pain effects, it misleads zero-shot language-model classifiers, and it has entered the published literature as a clinical finding. Correcting the documentation is simple, and the evidence and code to do so are public.

\section*{Data and code availability}
All evidence files, scripts, the literature-audit coding and the LLM-probe notebook are available at \url{https://github.com/tpattabhi/anginaflip}. An earlier version (18 Nov 2026), titled ``Typical Angina, Reversed: A Chest-Pain Coding Error in the UCI Heart Disease Data That Has Persisted Since 1990'', is archived at \url{https://doi.org/10.5281/zenodo.24358545}.

\section*{Acknowledgments and disclosures}
This work received no funding. The author reports no competing interests. The work was done independently and is not affiliated with any employer. The author used Claude (Anthropic) to write and run analysis code, retrieve and extract text from the literature, produce the primary literature-audit codes, and edit the manuscript. The author reviewed the primary codes and recoded a sample of papers; OpenAI Codex reviewed that recoding against the coding rules. The author reviewed, verified and takes responsibility for all content.

\bibliographystyle{unsrtnat}
\bibliography{refs}

\end{document}